\documentclass{svproc}

\usepackage[utf8]{inputenc}
\usepackage{graphicx}
\usepackage{hyperref}
\usepackage{multicol}
\usepackage{float}
\usepackage{listings}
\usepackage{amsmath}
\usepackage{amsfonts}
\usepackage{footmisc}
\usepackage{svg}

\title{A similarity measure of Gaussian process predictive distributions}
\author{Lucia Asencio-Mart\'in, Eduardo C. Garrido-Merch\'an}
\institute{Universidad Aut\'onoma de Madrid, Madrid, Spain\\
\email{lucia.asencio@estudiante.uam.es} \\ 
\email{eduardo.garrido@uam.es}}

\begin{document}

\maketitle

\begin{abstract}
 Some scenarios require the computation of a predictive distribution of a new value evaluated on an objective function conditioned on previous observations. We are interested on using a model that makes valid assumptions on the objective function whose values we are trying to predict. Some of these assumptions may be smoothness or stationarity. Gaussian process (GPs) are probabilistic models that can be interpreted as flexible distributions over functions. They encode the assumptions through covariance functions, making hypotheses about new data through a predictive distribution by being fitted to old observations. We can face the case where several GPs are used to model different objective functions. GPs are non-parametric models whose complexity is cubic on the number of observations. A measure that represents how similar is one GP predictive distribution with respect to another would be useful to stop using one GP when they are modelling functions of the same input space. We are really inferring that two objective functions are correlated, so one GP is enough to model both of them by performing a transformation of the prediction of the other function in case of inverse correlation. We show empirical evidence in a set of synthetic and benchmark experiments that GPs predictive distributions can be compared and that one of them is enough to predict two correlated functions in the same input space. This similarity metric could be extremely useful used to discard objectives in Bayesian many-objective optimization.
\end{abstract}
\section{Introduction}
Regression problems involve the prediction of the value $y$ associated with the evaluation of a point $\mathbf{x} \in \mathbb{R}^d$ in an objective function or unknown ground truth $f(\mathbf{x})$, where $d$ is the number of dimensions of $\mathbf{x}$ \cite{murphy2012machine} \cite{bishop2006pattern}. Let $\mathcal{X}$ be a subset of $\mathbb{R}^d$ called the input space. Supervised learning finds the values of the machine learning (ML) algorithm hyper-parameters $\boldsymbol{\theta}$ that make the algorithm calculate an accurate prediction of $f(\mathbf{x})$ via fitting the algorithm with a dataset $\mathcal{D} = \{(\mathbf{x}_i, y_i)| i = 1,...,n\}$, where $\mathbf{x}_i$ are points labelled with values $y_i$. ML algorithms then perform predictions $y^\star$ of new points $\mathbf{x}^\star$. If the ML algorithm does not compute an uncertainty $\sigma(\mathbf{x})$ of its predictions $y$, the user of the regression will not know the degree of certainty of the predictions done by the ML algorithm. This is the case of Deep neural networks \cite{lecun2015deep}, that do not provide uncertainty. Gaussian processes (GPs) \cite{mackay1998introduction} and Bayesian neural networks compute uncertainties of its predictions. A GP model is equivalent to a fully connected deep neural network with infinite number of hidden units in each layer \cite{lee2018deep}. GPs have successfully been used for regression problems where the uncertainty $\sigma(\mathbf{x})$ of the predictions $y$ is important \cite{rasmussen2003gaussian}.

GPs have also been used as probabilistic surrogate models in Bayesian optimization (BO) \cite{snoek2012practical} \cite{shahriari2015taking}. BO deals with the optimization of black-box functions. A black-box is a function whose analytical expression is unknown. Hence, its gradients are not accesible. It is very expensive to evaluate and the function evaluations are potentially noisy. The estimation of the generalization error of ML algorithms is considered to be a black-box function. We find other applications in structure learning of probabilistic graphical models \cite{cordoba2018bayesian} or even subjective tasks as suggesting better recipes \cite{garrido2018suggesting}. When not only one but several black-boxes are optimized, we deal with the Multi-objective BO scenario \cite{hernandez2016predictive}. If these objectives need to be optimized under the presence of constraints, we deal with the constrained multi-objective scenario \cite{garrido2019predictive}. BO suggest one point per evaluation, but it also can suggest several points in the constrained multi-objective scenario \cite{garrido2020parallel}. Typically, these problems involve the optimization of less than $4$ objectives. Many objective optimization has dealt with the optimization of more than $4$ objectives \cite{fleming2005many}. This scenario has not been targeted by BO. An approach to solve this scenario is to get rid of objectives that can be explained through the other objectives. In BO, these objectives are modelled by GPs. If we had a similarity measure of the predictive distribution computed by a GP over an input space $\mathcal{X}$, we could use it to propose an approach for the many objective BO scenario. This is precisely the motivation for this paper: proposing a specialist GP predictive distribution similarity metric to be used in the many objective BO scenario.

The purpose of our measure is to detect a GP that is so similar to another GP that we can stop fitting it in many objective BO. Hence, we cannot just apply the KL divergence, as it is just a similarity measure of probability distributions. Let $\boldsymbol{\theta}$ represent the set of parameters of a distribution $\mathcal{P}$. The KL divergence between two probability density functions $p(\boldsymbol{\theta})$ and $q(\boldsymbol{\theta})$ over continuous variables is given by the following expression:
\begin{equation}
\text{KL}(p(\boldsymbol{\theta})||q(\boldsymbol{\theta})) = \int_{-\infty}^{\infty} p(\boldsymbol{\theta})\text{log}(\frac{p(\boldsymbol{\theta})}{q(\boldsymbol{\theta})})d\boldsymbol{\theta}\,.
\end{equation}
As we can see, KL is not focused on things like the importance of the point and its neighbourhood that maximizes the objective function $f(\mathbf{x})$ but in all the probability distribution support. Our measure differs from KL divergence in the fact that we focus on particular characteristics of the GP predictive distribution that are relevant for discarding a GP in a hypothetical many objective BO scenario.

In this work, we focus on comparing GP predictive distributions, as it is the arguably most widely used model in BO \cite{snoek2012practical}. Nevertheless, our measure could also be applied to the predictive distributions of Bayesian neural networks or Random forests, widening its applicability. 

The paper is organized as follows. First, we briefly review the fundamental concepts of GPs. These concepts will be useful to better understand the purpose of our proposed measure. Then, we include a section describing our proposed similarity measure. We add empirical evidence of the practical use of this measure in an experiments section. Lastly, we illustrate conclusions about this work and further lines of research. 
\section{Gaussian Processes}
A Gaussian Process (GP) is a collection of random variables (of potentially infinite size), any finite number of which have (consistent) joint Gaussian distributions \cite{rasmussen2003gaussian}. We can also think of GPs as defining a distribution over functions where inference takes place directly in the space of functions \cite{rasmussen2003gaussian}. A GP can be used for regression of a function $f(\mathbf{x})$.

Let $\mathbf{X}=(\mathbf{x}_1,...\mathbf{x}_N)^T$ be the training matrix and $\mathbf{y}=(y_1,...,y_N)^T$ be a vector of labels to predict. We define as a dataset $\mathcal{D} = \{(\mathbf{x}_i, y_i)| i = 1,...,n\}$ the set of labeled instances. A GP is fully characterized by a zero mean and a covariance function $k(\mathbf{x},\mathbf{x}')$, that is, $f(\mathbf{x}) \sim \mathcal{G}\mathcal{P}(\mathbf{0},k(\mathbf{x},\mathbf{x}'))$. 

Given a set of observed data $\mathcal{D} = \{(\mathbf{x}_i, y_i)| i = 1,...,n\}$, where $y_i=f(\mathbf{x}_i) + \epsilon_i$ with $\epsilon_i$ some
additive Gaussian noise, a GP builds a predictive distribution for the potential values of $f(\mathbf{x})$ at a new input point $\mathbf{x}^\star$. This distribution is Gaussian. The GP mean, $\mu(\boldsymbol{x})$, is usually set to $0$. Namely,
$p(f(\mathbf{x}^\star)|\mathbf{y}) =\mathcal{N}(f(\mathbf{x}^\star)|
\mu(\mathbf{x}^\star),  v(\mathbf{x}^\star))$, where the mean $\mu(\mathbf{x}^\star)$ and variance $v(\mathbf{x}^\star)$ are respectively given by:
\begin{align}
\mu(\mathbf{x}^\star) & = \mathbf{k}_{\star}^{T} (\mathbf{K}+\sigma^{2}\mathbf{I})^{-1}\mathbf{y}\,, \\
v(\mathbf{x}^\star) & = k(\mathbf{x}_{\star},\mathbf{x}_{\star}) - \mathbf{k}_{\star}^T(\mathbf{K}+\sigma^{2} \mathbf{I})^{-1}\mathbf{k}_\star\,,
\label{eq:pred_dist}
\end{align}
where $\mathbf{y}=(y_1,\ldots,y_N)^\text{T}$ is a vector with the observations collected so far;
$\sigma^2$ is the variance of the additive Gaussian noise $\epsilon_i$;
$\mathbf{k}_\star = \mathbf{k}(\mathbf{x}_*)$ is a $N$-dimensional vector with the prior covariances between the test point $f(\mathbf{x}^\star)$ and
each of the training points $f(\mathbf{x}_i)$; and $\mathbf{K}$ is a $N\times N$ matrix with the prior covariances
among each $f(\mathbf{x}_i)$, for $i=1,\ldots,N$. Each element $\mathbf{K}_ij = k(\mathbf{x}_i, \mathbf{x}_j)$ of the matrix $\mathbf{K}$ is given by the covariance function between each of the training points $\mathbf{x}_i$ and $\mathbf{x}_j$ where $i,j = 1,...,N$ and $N$ is the total number of training points. The particular characteristics assumed for $f(\mathbf{x})$ (\emph{e.g.}, level of smoothness, additive noise, etc.) are specified by the covariance function $k(\mathbf{x},\mathbf{x}')$ of the GP. A popular example of covariance function is the squared exponential, given by:
\begin{align}
k(\mathbf{x},\mathbf{x}') & = \sigma^2_f \exp\left(-\frac{r^2}{2\ell^2}\right) + \sigma^2_n\delta_{pq}\,,
\end{align}
where $r$ is the Euclidean distance between $\mathbf{x}$ and $\mathbf{x}'$, $\ell$ is a hyper-parameter known as length-scale that controls the smoothness of the functions generated from the GP, $\sigma^2_f$ is the amplitude parameter or signal variance that controls the range of variability of the GP samples and $\sigma^2_n\delta_{pq}$ is the noise variance that applies when the covariance function is computed to the same point $k(\mathbf{x},\mathbf{x})$. Those are hyper-parameters of the GP. Let $\boldsymbol{\theta}$ be the set of all those hyper-parameters. We can find point estimates for the hyper-parameters $\boldsymbol{\theta}$ of the GP via optimizing the log marginal likelihood. The marginal likelihood is given by the following expression:
\begin{equation}
\log p(\mathbf{y}|\mathcal{X}, \boldsymbol{\theta}) = - \frac{1}{2} \mathbf{y}^T (\mathbf{K} + \sigma_n^{2} I)^{-1} \mathbf{y} - \frac{1}{2} \log | \mathbf{K} + \sigma_n^2I| - \frac{n}{2} \log 2\pi\,.
\end{equation}
The previous analytical expression can be optimized to obtain a point estimate $\boldsymbol{\theta}^\star$ for the hyper-parameters $\boldsymbol{\theta}$. We can optimize it through a local optimizer such as L-BFGS-B \cite{zhu1997algorithm} and via the analytical expression of the marginal likelihood gradient $\nabla_{\boldsymbol{\theta}} \log (\mathbf{y}|\mathbf{X}, \boldsymbol{\theta}) = (\partial \log (\mathbf{y}|\mathbf{X}, \boldsymbol{\theta} /\partial \theta_1 , ..., \partial \log (\mathbf{y}|\mathbf{X}, \boldsymbol{\theta}) / \partial \theta_M )^T$ whose partial derivatives are given by:
\begin{align}
\frac{\partial}{\partial \theta_j} \log (\mathbf{y}|\mathbf{X}, \boldsymbol{\theta}) & = \frac{1}{2} \mathbf{y}^T \mathbf{K}^{-1} \frac{\partial \mathbf{K}}{\partial \theta_j} \mathbf{K}^{-1} \mathbf{y} - \frac{1}{2}\text{tr}(\mathbf{K}^{-1}\frac{\partial \mathbf{K}}{\partial \theta_j}) \nonumber \\
& = \frac{1}{2} \text{tr} ((\boldsymbol{\alpha}\boldsymbol{\alpha}^{T} - \mathbf{K}^{-1})\frac{\partial \mathbf{K}}{\partial \theta_j}) \,,
\end{align}
where $\boldsymbol{\alpha} = \boldsymbol{K}^{-1}\mathbf{y}$ and $M$ is the number of hyper-parameters.

\section{The Similarity Measure}

Let $f(\textbf{x})$, $g(\textbf{x})$ be two GPs. Let us work under the assumption that their covariance functions $k(\mathbf{x}, \mathbf{x})$ have similar analytical expressions (e.g., both are squared exponential functions). Given a set $\mathbf{X}_{\star}$ of input points, we can compute $\mu_f(\textbf{X}_{\star})$ and $\mu_g(\textbf{X}_{\star})$, the predicted mean vectors for each process, as well as $v_f(\textbf{X}_{\star}(\textbf{X}_{\star})$ and $v_g(\textbf{X}_{\star})$, their covariance matrices.\\
We now define a notion of distance between these two processes, firstly  presenting its mathematical expression:
\begin{align} 
d(f(\textbf{x}) , g(\textbf{x})) =&  \varepsilon_1d_1\left(T(\mu_f(\textbf{X}_{\star})), \mu_g(\textbf{X}_{\star}), \delta\right) + \nonumber \\ 
&  \varepsilon_2d_2\left(v_f(\textbf{X}_{\star}), v_g(\textbf{X}_{\star})\right) + \nonumber \\
&  \left( 1 - \varepsilon_1 - \varepsilon_2\right)\left(1 - \rho\left(\mu_f(\textbf{X}_{\star}), \mu_g(\textbf{X}_{\star})\right)\right) \,.
\end{align}
As it can be seen, the measure is given following a weighted sum model (WSM \cite{TriantaphyllouEvangelos2000Mdmm}). The WSM contains three components to which we will refer as:
\begin{align}
& s_1 = \varepsilon_1d_1\left(T(\mu_f(\textbf{X}_{\star})), \mu_g(\textbf{X}_{\star}), \delta\right) \,, \nonumber \\
& s_2 = \varepsilon_2d_2\left(v_f(\textbf{X}_{\star}), v_g(\textbf{X}_{\star})\right) \,, \nonumber \\
& s_3 = \left( 1 - \varepsilon_1 - \varepsilon_2\right)\left(1 - \rho\left(\mu_f(\textbf{X}_{\star}), \mu_g(\textbf{X}_{\star})\right)\right) \,.
\end{align}
The objective of $s_1$ and $s_3$ is to describe the distance between both mean vectors, while $s_2$ aims to reflect the distance between the covariance matrices. We will now analyze each of these components and their respective parameters.\\ \\
The first one, $s_1$, is given in terms of a tolerance $\delta$, a transformation function $T$ and a distance function $d_1$ between the mean vectors.\\
The election of $T$ describes when two  mean vectors $\mu_f(\textbf{X}_{\star}) \neq \mu_g(\textbf{X}_{\star})$ should be considered equal. For example, since we will be optimizing these vectors, if $\mu_f(\textbf{X}_{\star}) = 2\mu_g(\textbf{X}_{\star})$, their critical points will be the exact same and we might want to consider them as a single vector. Although $T$ can be chosen by the user depending on their needs, the proposed implementation provides a function $T(\mu_f(\textbf{X}_{\star})) = a\mu_f(\textbf{X}_{\star})+ b\textbf{1}$, where $a>0$ and $b$ are scalars chosen with the least squares method to give the best fit of  $\mu_f(\textbf{X}_{\star})$ onto $\mu_g(\textbf{X}_{\star})$. This transformation reflects the fact that two vectors that are proportional and whose difference is constant behave in the same way in terms of optimization.\\
With function $d_1(\cdot)$, the user is able to choose in which way they want to measure the distance between $T(\mu_f(\textbf{X}_{\star}))$ and $\mu_g(\textbf{X}_{\star})$. Several options are given to the user in our implementation, each of them being convenient depending on the nature of the problem modeled by the GP. Some of these options are to define $d_1$ as the number (or percentage) of points where $T(\mu_f(\textbf{X}_{\star}))\neq\mu_g(\textbf{X}_{\star}) $, or as a $p$-norm ($d_1= \|{\mu_g(\textbf{X}_{\star}) - T(\mu_f(\textbf{X}_{\star}))}\|_{p} $) which includes euclidean norm, infinity norm, etc.\\
Lastly, with $\delta$ the user is allowed to change the desired level of tolerance given to $d_1$, i.e. the distance is calculated only among the vectors' elements where the chosen $d_1$ is greater than $\delta$ in its element-wise operations. For example, if we chose a $1$-norm as $d_1$, $s_1$ would be computed as $\sum_{|\mu_g - T(\mu_f)| > \delta}{|\mu_g(\textbf{X}_{\star}) - T(\mu_f(\textbf{X}_{\star}))|}$.\\\\
The other weighted sum term used to compare the two mean vectors is $s_3$. It is the only fixed term in the sum, and it represents the Pearson correlation coefficient between the GP means. \\
This coefficient is defined as 
\begin{equation}
{\rho(\mu_f(\textbf{X}_{\star}),\mu_g(\textbf{X}_{\star}))={\frac {\mathbb{E} [(\mu_f(\textbf{X}_{\star})-\overline{\mu _f(\textbf{X}_{\star})})(\mu_g(\textbf{X}_{\star})-\overline{\mu _y(\textbf{X}_{\star})}))]}{\sigma_{\mu_f}\sigma _{\mu_g}}}} \,,
\label{eq:pearson}
\end{equation}
where $\overline{\mu(\cdot)}$ denotes the mean value of a vector $\mu$ and $\sigma_\mu$ is its standard deviation.\\The reason we were first interested in this operator is because of its interpretation. The coefficient $\rho(\mu_f(\textbf{X}_{\star}), \mu_g(\textbf{X}_{\star}))$ ranges from -1 to 1. If it equals 1, there is a (positive) linear equation describing $\mu_g(\textbf{X}_{\star})$ in terms of $\mu_f(\textbf{X}_{\star})$; if it equals -1, this linear equation has a negative slope and, when it is 0, no linear correlation between $\mu_f(\textbf{X}_{\star})$ and $\mu_g(\textbf{X}_{\star})$ exists. \\
Moreover, following Eq. (\ref{eq:pearson}), $\rho$ increases whenever $\mu_f(\textbf{X}_{\star})$ and $\mu_g(\textbf{X}_{\star})$ both increase or decrease. It decreases when their growth behaviour is different. This is very valuable for our problem, since we need to identify whether two vectors are increasing and decreasing in a similar fashion, i.e., their maximums and minimums lie around the same positions.\\
We found this to be the most accurate way of detecting similar processes, since it detects that sample vectors of functions like $x^6$ and $x^2$, which would a priori seem very different using any conventional vector distance (one grows much faster than the other) behave essentially the same: they both decrease from $-\infty$ to 0, have a minimum in 0 and increase towards $\infty$.
\\\\
Lastly, the addend $s_2$ is intended to measure the distance between both predictive variances, and therefore any matrix norm could be used for this purpose. \\
We have not found the matrices distance to be significant when it comes to deciding whether two GPs should be optimized analogously, although this might be because of working under the assumption that $f(\textbf{X}_{\star})$ and $g(\textbf{X}_{\star})$ have similar covariance functions.  In  case the user wants to make use of the matrices similarity, note that entrywise matrix norms should be preferred over the ones induced by vector norms because of their lower computational cost \cite{0908.1397}.\\ 
As future work, these matrices could be used to measure the uncertainty of the distance between two GPs, since they represent the uncertainty of the GPs' predictions.
\section{Experiments}
\begin{figure}[t]
\includegraphics[width=\textwidth]{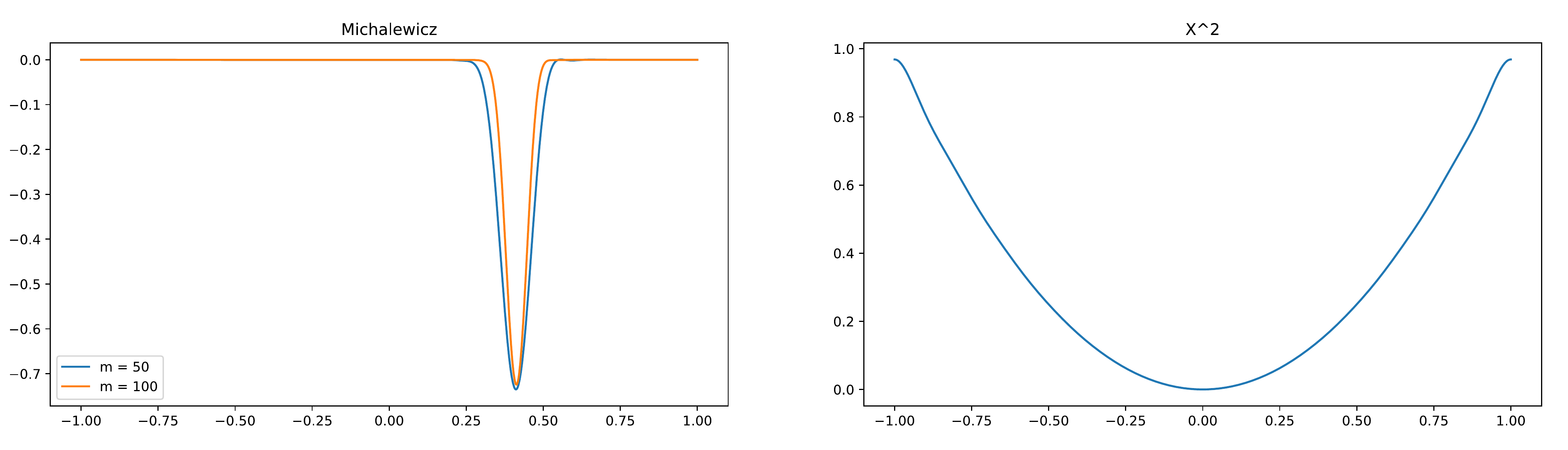}
\centering
\caption{Toy functions for one dimension}
\label{fig:1d}
\end{figure}

\begin{figure}[t]
\includegraphics[width=\textwidth]{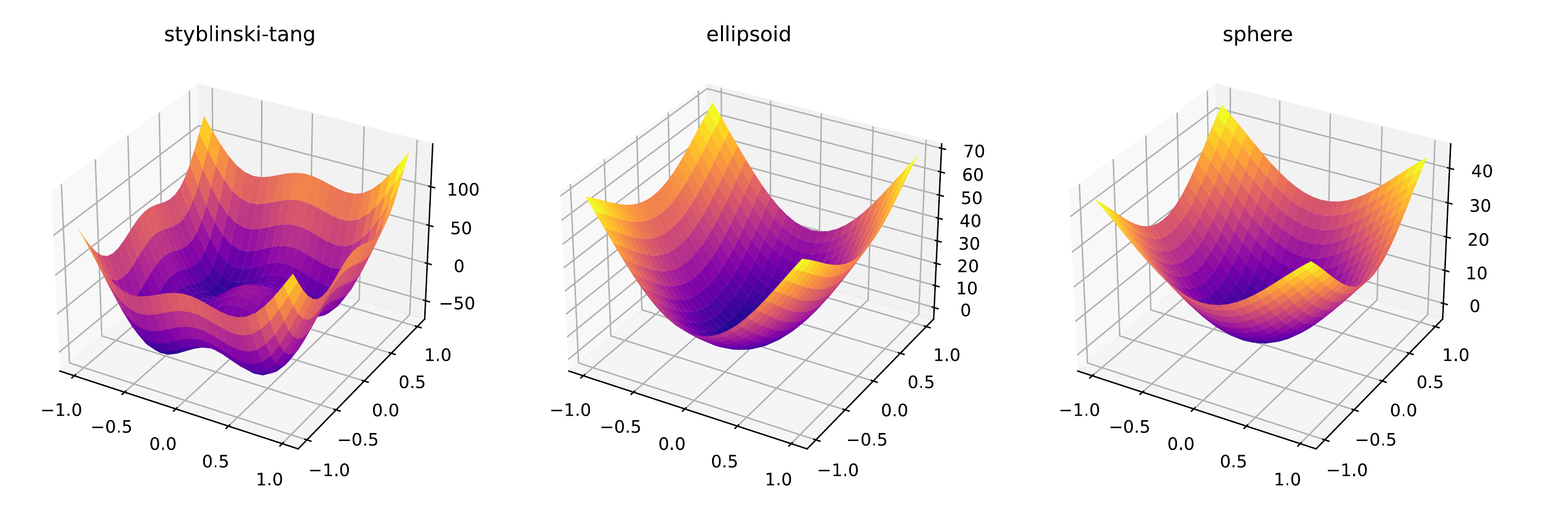}
\centering
\caption{Bowl-shaped functions}
\label{fig:bowl}
\end{figure}

\begin{figure}[t]
\includegraphics[width=\textwidth]{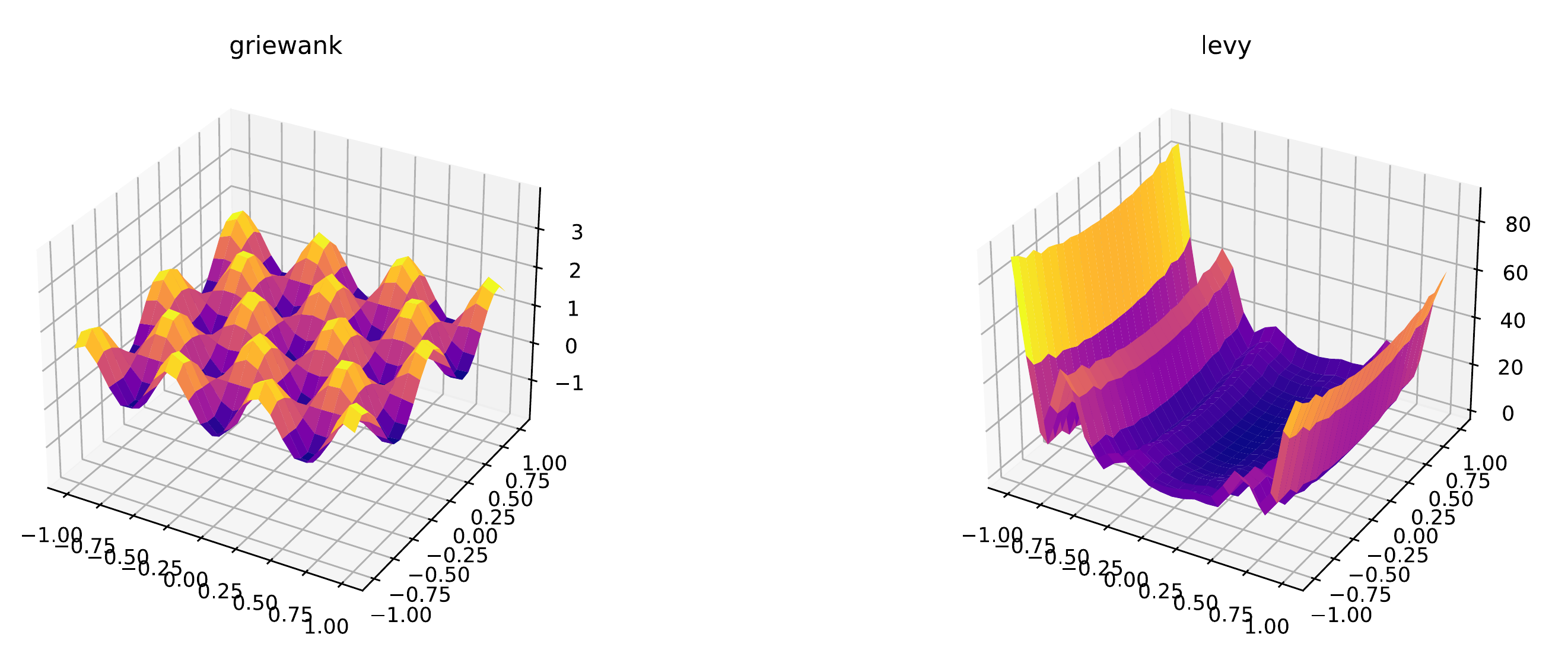}
\centering
\caption{Griewank and Levy functions}
\label{fig:misc}
\end{figure}

\begin{figure}[t]
\includegraphics[width=\textwidth]{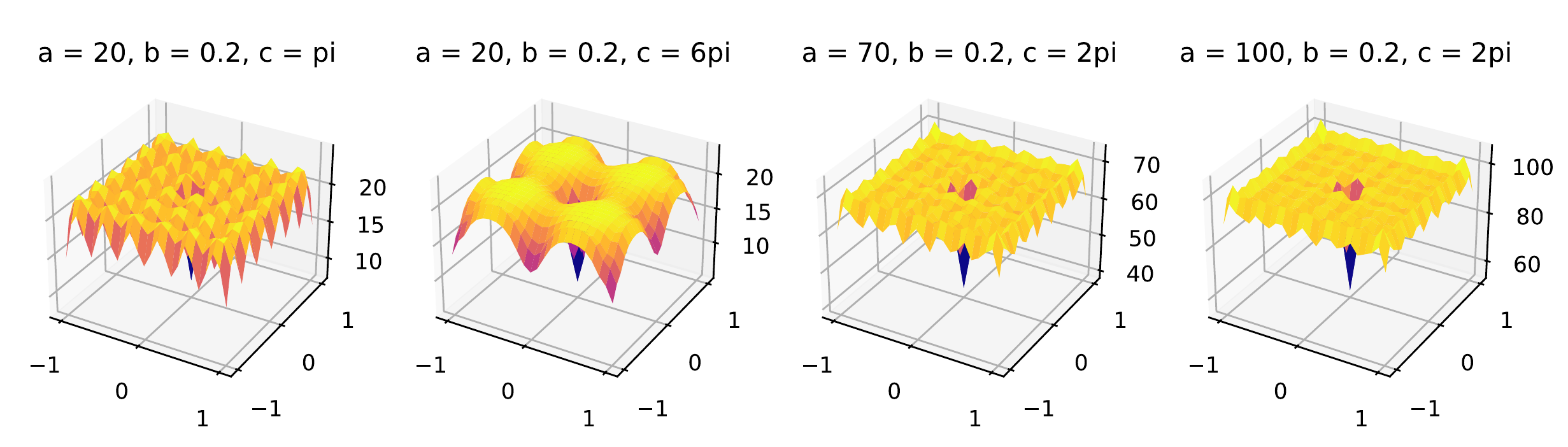}
\centering
\caption{Ackley functions}
\label{fig:ackley}
\end{figure}

For all the experiments, we have set the parameter $\delta=0$ and, since we didn't found the covariance matrices distance to be significant under our hypothesis, we used $\varepsilon_1 = 0.25$ and $\varepsilon_2 = 0$ (therefore $\rho$'s weight is $0.75$). For the transformation $T$, the previously explained linear transformation using the least squares method was used. We have chosen $d_1$ to be the average relative distance between the points in the two mean vectors, i.e., the mean of the vector given by $|T\left(\mu_f(\textbf{X}_\star)\right) - \mu_g(\textbf{X}_\star)|$ divided by the subtraction of the greatest element found in the two vectors minus the smallest.  \\For the following examples, various GPs were fitted taking sample vectors from benchmark functions. We will now discuss some of the results obtained by applying our measure to find the distance between them.\\\\
We will start with some uni-dimensional toy functions. We have chosen to compare three GPs from which we know that two of them are very similar and that the third one behaves differently from the other two. A plot of their mean vectors can be seen in Fig.\ref{fig:1d}.
The first one models a Michalewictz function (defines as $f_1(x)=-\sin{x}\left(\sin{\frac{x^2}{\pi}}\right)^{2m}$) with parameter $m=50$, the second one a Michalewictz function with $m = 100$ and the third one models a parabola $x^2$.
The correlation between the predicted means of the two Michalwicz GPs is $0.97$, and the average relative distance between them is $0.02$. In total, the distance calculated by our measure is $0.02$ over $1$, i.e., these GPs are very similar according to our function. 
On the other hand, when comparing the Michalewictz GP that has $m=100$ with the parabola, we obtain a correlation of $0.12$, which reflects how different their growth behaviour is, and an average relative distance of $0.27$. In total a distance of $0.72$ over 1.\\\\
We now compare three 2D bowl-shaped processes: one of them modelling a Styblinski-Tang function, another one an ellipsoid and the third one a sphere. \\The Syblinski-Tang function is given by $f_2(\textbf{x}) = \frac{1}{2}\sum_{i=1}^2\left(x_i^4-16x_i^2+5x_i\right)$, the ellipsoid is
$f_3(\textbf{x}) = \sum_{i=1}^2\sum_{j=1}^ix_j^2$ and the sphere function is $f_4(\textbf{x})= x_1^2+x_2^2$
Since the three are bowl shaped, they are somewhat similar, but from Fig.\ref{fig:bowl} we can tell that Styblinski-Tang is slightly different from the others. Indeed, when we compare the sphere and the ellipsoid processes we obtain a $0.94$ correlation between  their predicted mean vectors, and 0.06 is their average relative distance. Overall, the distance is $0.05$ which means these two functions are very similar. In contrast, when we compare the ellipsoid with the Styblinski-Tang process, we obtain a correlation of $0.74$: a value which is close to 1, reflecting the fact that both shapes are similar in a way, but not as close as in the previous comparison because the Styblinski-Tang bowl is different from the ellipsoid. The average relative distance is 0.13 (again a value which is greater than before) and the total distance of $0.22$ over 1, which shows that these processes are similar, but not as much as the previous ones.\\\\
We will now compare two processes modelling two functions that we know are not similar, illustrated in Fig.\ref{fig:misc}. The first one is a Griewank function $f_5(\textbf{x}) = \frac{1}{4000}\left(x_1^2+x_2^2\right)-\cos
(x_1)\cos\left(\frac{x_2}{\sqrt{2}}\right)+1$, and the second one a levy function $f_6(\textbf{x}) = \sin^2(\pi w_1) + (w_1-1)^2\left(1+10\sin^2(\pi w_1+1)\right)+ (w_2-1)^2\left(1+\sin^2(2\pi w_2)\right)$ where $w_i = 1 + (x_i-1)\frac{1}{4}$\\
Because of how different their shapes are, the correlation between the predicted mean is 0.04, almost no correlation at all. Their average relative distance is 0.16, and the distance between the processes is 0.75 over 1.\\\\
Finally, we will be comparing some Ackley functions. Recall that an ackley function depending on parameters $a$, $b$ and $c$ is given by 
$$f_7(\textbf{x)}=  -a\exp \left( -b\sqrt{\frac{1}{d}\sum_{i = 1}^2x_i^2}\right) 
-\exp \left( -\frac{1}{2}\sum_{i = 1}^2\cos\left(cx_i\right)\right) + a + e$$
We first compare two processes with fixed $a = 20$ and $b = 0.2$. The parameter $c$ equals $\pi$ for one process and $6\pi$ for the other, which results in unsimilar predicted means plots as can be seen in Fig. \ref{fig:ackley}. The correlation between them is 0.31, and the average relative distance 0.15, giving an overall distance of 0.55 over 1. Lastly, we compare two Ackley processes which, despite having different $a$ values, have a very similar shape (Fig.\ref{fig:ackley}). We chose $b=0.2, c=2\pi$ and $a=70$ for one process and $a=100$ for the other. Their correlation is very close to 1, 0.98, and their average relative distance is 0.01. A total distance of 0.01, which means these processes are indeed very similar in their shape.
\section{Conclusions}
In this paper, we have proposed a similarity metric for GP predictive distributions in order to be used in a many objective BO scenario \cite{ishibuchi2008evolutionary}. Thanks to this metric, we are able to measure how similar are the predictions given by two GP predictive distributions. We have illustrated the results of this measure in a set of synthetic and benchmark functions. This measure will be used in constrained multi-objective BO, in scenarios with more than $3$ objectives. Another possible use of this metric is its application for the cognitive architectures of brain inspired autonomous robots \cite{garrido2020artificial}. In this architecture, every objective can model a particular emotion, that conditions the policy of the robot. The robot wants to simultaneously optimize conflicting emotions like happiness, social contacts or being full of energy. As human beings are only phenomenally conscious of a small set of emotions in every moment, we can imitate that behavior by discarding redundant emotions through this metric and making the global workspace only aware of the most relevant emotions \cite{soto2017artificial} \cite{merchan2020machine}. 

\section*{Acknowledgments}

The authors gratefully acknowledge the use of the facilities of Centro
de Computaci\'on Cient\'ifica (CCC) at Universidad Aut\'onoma de
Madrid. The authors also acknowledge financial support from Spanish
Plan Nacional I+D+i, grants TIN2016-76406-P and TEC2016-81900-REDT.

\bibliographystyle{plain}
\bibliography{main}
\end{document}